# Topeax - An Improved Clustering Topic Model with Density Peak Detection and Lexical-Semantic Term Importance

Márton Kardos
Aarhus University
Student no. 202105399
martonkardos@cas.au.dk

## Abstract

Text clustering is today the most popular paradigm for topic modelling, both in academia and industry. Despite clustering topic models' apparent success, we identify a number of issues in Top2Vec and BERTopic, which remain largely unsolved. Firstly, these approaches are unreliable at discovering natural clusters in corpora, due to extreme sensitivity to sample size and hyperparameters, the default values of which result in suboptimal behaviour. Secondly, when estimating term importance, BERTopic ignores the semantic distance of keywords to topic vectors, while Top2Vec ignores word counts in the corpus. This results in, on the one hand, less coherent topics due to the presence of stop words and junk words, and lack of variety and trust on the other. In this paper, I introduce a new approach, Topeax, which discovers the number of clusters from peaks in density estimates, and combines lexical and semantic indices of term importance to gain high-quality topic keywords. Topeax is demonstrated to be better at both cluster recovery and cluster description than Top2Vec and BERTopic, while also exhibiting less erratic behaviour in response to changing sample size and hyperparameters.

## 1. Introduction

Topic models are statistical models that can identify latent topic variables in a selection of documents, and can describe them with keywords/phrases (Blei, 2012). Older topic models typically relied on bag-of-words representations of text, and conceptualized topic discovery as recovering latent factors that generate word content in documents. Due to advances in text embedding, documents can now be encoded into dense neural representations (Le & Mikolov, 2014; Reimers & Gurevych, 2019).

### 1.1. Clustering Topic Models

Neural text embeddings are easier to cluster than bag-of-words document vectors, and this has allowed researchers to conceptualize topic modelling as discovering clusters of documents in embedding space.

The Top2Vec model (Angelov, 2020) relies on a multi-stage pipeline for discovering interpretable topics in embedding spaces. Document embeddings are first reduced to a lower dimensionality using a manifold learning technique called UMAP (Healy & McInnes, 2024). Next, documents are clustered using a density-based technique called HDBSCAN (Campello et al., 2013), which in theory, can also determine the number of clusters empirically. After discovering clusters, Top2Vec assigns importance to words based on their proximity in embedding space to topic vectors, which are centroids of the discovered clusters. More recently, Angelov & Inkpen (2024) have used a sliding window over BERT embeddings to get clusters of contextualized document chunks, introducing c-Top2Vec. The clustering methodology and term-importance estimation schemes, however, remain the same.

$$t_k = \frac{\sum_{d \in T_k} x_d}{|T_k|}; \beta_{kj} = \cos(t_k, w_j)$$

where $t_k$ is the embedding of topic $k$ and $x_d$ is the embedding of document $d$, $T_k$ is the set of documents in topic $k$, $w_j$ is the embedding of term $j$ and $\beta_{kj}$ is the importance of term $j$ in topic $k$.

BERTopic (Grootendorst, 2022) is a very similar model, with the only difference being that it uses a weighting scheme called c-TF-IDF (see Appendix B for formula) for computing term importance instead. This approach is more theoretically correct, as Top2Vec makes the assumption that clusters are spherical, which is likely not the case with a density-based clustering model.

Users will commonly find that these topic models discover a larger number of topics than they find useful for their analysis. In order to combat this, both methods have a *hierarchical topic reduction* method. In both cases, users have to specify how many topics they would like to have in the end, and then clusters are merged until this desired number is reached. BERTopic utilizes agglomerative clustering with average linkage, while Top2Vec merges the smallest cluster to the closest one based on centroid proximity.

Although the clustering topic models have been enjoying popularity in academia (BERTopic has at the time of writing 3726 citations on Google Scholar, while Top2Vec has 827), they are plagued by a number of problems. Hoyle et al. (2025) found, using extensive human evaluation, BERTopic no better than older models like LDA. Kardos et al. (2025b) found that BERTopic often includes stop-words in topics, and Top2Vec is usually negatively affected by higher-dimensional embeddings. In addition, these models have usually been evaluated in a setting where the topics were reduced hierarchically, including the original papers. Our knowledge is limited on how well these models perform when they have to determine the number of topics themselves, and how well they recover clusters in corpora.

This papers' contributions can be summarized as follows: Firstly, I evaluate these models in a free-clustering scenario, without specifying the number of clusters, both on how well they match gold cluster labels, as well as topic quality. Secondly, they are evaluated based on their sensitivity to subsampling and hyperparameters. And thirdly, I introduce a novel method, Topeax, which outperforms Top2Vec and BERTopic on these tasks.

## 2. Model Specification

I introduce Topeax, a novel topic modelling approach based on document clustering. The model is implemented in the Turftopic Python package (Kardos et al., 2025a), following scikit-learn API conventions (Buitinck et al., 2013). Example usage, along with figure and keywords are presented in Appendix A. This section will outline how Topeax discovers topics.

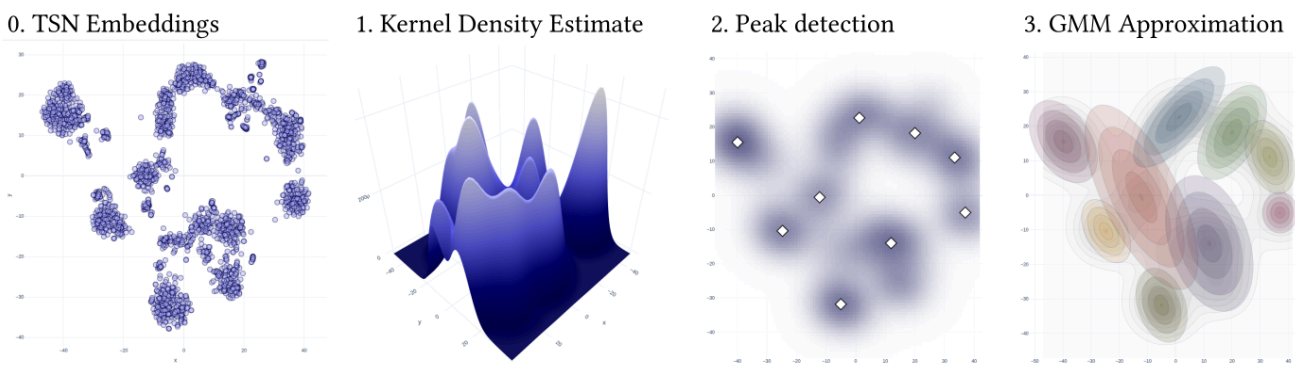

Figure 1: A schematic overview of the Peax clustering algorithm.
Illustrations were generated from the *political ideologies dataset*[1].

---

[1] https://huggingface.co/datasets/JyotiNayak/political_ideologies

## 2.1. Dimensionality Reduction

Unlike other clustering topic models, Topeax relies on t-Distributed Stochastic Neighbour Embeddings (Maaten & Hinton, 2008) instead of UMAP. Cosine distance is used as the distance metric for dimensionality reduction, due to its wide-spread use in model training and downstream applications. The number of dimensions was fixed to 2 in all experiments, as this allows for easier visualization. While it has been demonstrated that TSNE can be sensitive the chosen value of `perplexity` (Wattenberg et al., 2016), we will show that, within a reasonable range, this will not have an effect on the number of topics or topic quality.

## 2.2. The Peax Clustering Model

For the clustering step in the modelling pipeline, I introduce a new technique for clustering, termed **Peax**, which clusters documents based on density peaks in the reduced document space.

The Peax algorithm consists of the following steps:

1. A Gaussian Kernel Density Estimate (KDE) is obtained over the reduced embeddings. Bandwidth is determined with the Scott method (Scott, 1979).
2. The KDE is evaluated on a 100x100 grid over the range of the embeddings.
3. Density peaks are detected by applying a local-maximum filter to the KDE heatmap. A neighbourhood connectivity of 25 is used, which means, every pixel is included within a 5 unit radius.
4. Cluster centres are assigned to these density peaks. The density structure of each cluster is estimated by fitting a Gaussian mixture model, with its means fixed to the peaks, using the Expectation-Maximization algorithm. Documents are assigned to the component with the highest responsibility:

$$\hat{z}_d = \arg\max_k(r_{kd}) \text{ , and } r_{kd} = p(z_k = 1 \mid \hat{x}_d)$$

where $\hat{z}_d$ is the estimated underlying component assigned to document $d$, $\hat{x}_d$ is the TSN embedding of document $d$, and $r_{kd}$ is the responsibility of component $k$ for document $d$.

## 2.3. Term Importance Estimation

To mitigate the issues experienced with c-TF-IDF and centroid-based term importance estimation in previously proposed clustering topic models, I introduce a novel approach that uses a combination of a embedding-based and a lexical term importance.

### 2.3.1. Semantic Importance

Semantic term importance is estimated similar to Angelov (2020), but, since we have access to a probabilistic, non-spherical model, and cluster boundaries are not hard, topic vectors are estimated from the responsibility-weighted average of document embeddings.

$$t_k = \frac{\sum_d r_{kd} \cdot x_d}{\sum_d r_{kd}}; s_{kj} = \cos(t_k, w_j)$$

where $t_k$ is the embedding of topic $k$ and $x_d$ is the embedding of document $d$, $w_j$ is the embedding of term $j$ and the semantic importance of term $j$ for cluster $k$ is $s_{kj}$.

### 2.3.2. Lexical Importance

Instead of relying on a tf-idf-based measure for computing the valence of a term in a corpus, an information-theoretical approach is used. I estimate the lexical importance of a term for a cluster by computing the mutual information of the term's presence with the cluster label. Due to its easier interpretability, I opt for using normalized pointwise mutual information (NPMI), which has been historically used for collocation extraction (Bouma, 2009) and topic-coherence evaluation (Röder et al., 2015). Pointwise mutual information is calculated by taking the logarithm of the fraction of conditional and marginal word probabilities:

$$\text{pmi}_{kj} = \log_2 \frac{p(v_j|z_k)}{p(v_j)}$$

where $p(v_j|z_k)$ is the conditional probability of word $j$ given the presence of topic $z$, and $p(v_j)$ is the probability of word $j$ occurring.

A naive approach might include estimating these probabilities empirically:

$$p(v_j) = \frac{n_j}{\sum_i n_i}, \text{ and } p(v_j \mid z_k) = \frac{n_{jt}}{\sum_i n_{it}}$$

where $n_j$ is the number of times word $j$ occurs, $n_{jt}$ is the number of times word $j$ occurs in cluster $t$.

This would, however, overestimate the importance of rare words in the clusters where they appear. We can therefore opt for a mean-a-posteriori estimate under a symmetric dirichlet prior with an $\alpha$ *smoothing* parameter, which is analyticaly tractable:

$$p(v_j) = \frac{n_j + \alpha}{N\alpha + \sum_i n_i}, \text{ and } p(v_j \mid z_k) = \frac{n_{jt} + \alpha}{N\alpha + \sum_i n_{it}}$$

where $N$ is the size of the vocabulary. In further analysis, $\alpha = 2$ will be used. Since regular PMI scores have no lower bound, we normalize them to obtain NPMI:

$$\text{npmi}_{kj} = \frac{\text{pmi}_{kj}}{-\log_2 p(v_j, z_k)}, \text{ where } p(v_j, z_k) = p(v_j|z_k) \cdot p(z_k)$$

### 2.3.3. Combined Term Importance

To balance the semantic proximity of keywords to topic embeddings and cluster-term occurrences, a combined approach is used, which consists in the geometric mean of min-max normalized lexical and semantic importance:

$$\beta_{kj} = \sqrt{\frac{1+\text{npmi}_{kj}}{2} \cdot \frac{1+s_{kj}}{2}}$$

## 3. Experimental Methods

Since one of the main strengths of clustering approaches is that they can supposedly find the number of clusters in the data, a good clustering topic model should roughly align with a human clustering of the data, and should be able to describe these clusters effectively. In the following section, I outline the benchmark used to evaluate models on these aspects.

Reproducible scripts used for evaluation, along with instructions on how to run them, are made available in the `x-tabdeveloping/topeax-eval`[2] Github repository. Results for all evaluations can be found in the `results/` directory.

### 3.1. Datasets

In order to evaluate models on a variety of domains, I used openly available datasets with gold labels. This included all clustering tasks from the new version of the Massive Text Embedding Benchmark `MTEB(eng, v2)` (Enevoldsen et al., 2025). To avoid evaluating on the same corpus twice, the P2P variants of the tasks where used. In addition an annotated Twitter topic-classification dataset (Kim et al., 2012), and a BBC News dataset (Greene & Cunningham, 2006) was used. I report descriptive statistics in Appendix C.

### 3.2. Models

To compare Topeax with existing approaches, it was run on all corpora alongside BERTopic and Top2Vec. Implementations were sourced from the Turftopic (Kardos et al., 2025a) Python package.

---
[2]https://github.com/x-tabdeveloping/topeax-eval

For the main analysis, default hyperparameters were used from the original BERTopic and Top2Vec packages respectively, as these give different clusterings. All models were run with both the `all-MiniLM-L6-v2`, and the slightly larger `all-mpnet-base-v2` sentence encoders (Reimers & Gurevych, 2019), as well as Google's `embeddinggemma-300m` (Schechter Vera et al., 2025) to control for embedding size and quality. The models were fitted without filtering for stop words and uncommon terms, since state-of-the art topic models have been shown to be able to handle such information without issues (Kardos et al., 2025b).

### 3.3. Metrics

For evaluating model performance, both clustering quality and topic quality was evaluated. I evaluated the faithfulness of the predicted clustering to the gold labels using the Fowlkes-Mallows index (Fowlkes & Mallows, 1983). The FMI, is very similar to the F1 score for classification, in that it also intends to balance precision and recall. Unlike F1, however, FMI uses the geometric mean of these quantities:

$$\text{FMI} = \frac{N_{\text{TP}}}{\sqrt{(N_{\text{TP}}+N_{\text{FP}}) \cdot (N_{\text{TP}}+N_{\text{FN}})}}$$

where $N_{\text{TP}}$ is the number of pairs of points that get clustered together in both clusterings (true positives), $N_{\text{FP}}$ is the number of pairs that get clustered together in the predicted clustering but not in the gold labels (false positives) and $N_{\text{FN}}$ is the number of pairs that do not get clustered together in the predicted clustering, despite them belonging together in the gold labels (false negatives).

For topic quality, I adopt the methodology of Kardos et al. (2025b), with minor differences. I use GloVe embeddings (Pennington et al., 2014) for evaluating internal word embedding coherence instead of Skip-gram. As such, topic quality was evaluated on topic diversity $d$, external word embedding coherence $C_{\text{ex}}$ using the `word2vec-google-news-300` word embedding model, as well as internal word embedding coherence $C_{\text{in}}$ with a GloVe model trained on each corpus. Ideally a model should both have high intrinsic and extrinsic coherence, and thus an aggregate measure of coherence can give a better estimate of topic quality: $\bar{C} = \sqrt{C_{\text{in}} \cdot C_{\text{ex}}}$. In addition an aggregate metric of topic quality can be calculated by taking the geometric mean of coherence and diversity $I = \sqrt{\bar{C} \cdot d}$. We will also refer to this quantity as *interpretability*.

### 3.4. Sensitivity to Perplexity

Both TSNE and UMAP, have a hyperparameter that determines, how many neighbours of a given point are considered when generating lower-dimensional projections, this hyperparameter is usually referred to as *perplexity*. It is also known that both methods are sensitive to the choice of hyperparameters, and depending on these, structures, that do not exist in the higher-dimensional feature space might appear in the lower-dimensional representations (Coenen & Pearce, n.d.; Wattenberg et al., 2016). In order to see how this affects the Topeax algorithm, in comparison with other clustering topic models, I fit each model to the 20 Newsgroups corpus from `scikit-learn`, using `all-MiniLM-L6-v2` with `perplexities=[2, 5, 30, 50, 100]`. This choice of values was inspired by Wattenberg et al. (2016). Each model was evaluated on the metrics outlined above.

### 3.5. Subsampling Invariance

Ideally, a good topic model should roughly recover the same topics, and same number of topics in a corpus even when we only have access to a subsample of that corpus, assuming that the underlying categories are the same. We should also expect that a model having access to the full corpus, instead of a subsample, should yield higher quality results. I fit each model on the same corpus and embeddings as in the perplexity sensitivity test, and evaluate them on the previously outlined metrics. Subsample sizes are the following: `[250, 1000, 5000, 10_000, "full"]`.

# 4. Results

Topeax substantially outperformed both Top2Vec and BERTopic in cluster recovery, as well as the quality of the topic keywords (see Figure 2). A regression analysis predicting Fowlkes-Mallows index from model type, with random effects and intercepts for encoders and datasets was conducted. The regression was significant at $\alpha = 0.05$. ($R^2 = 0.127$, $F = 4.368$, $p = 0.0169$). Both BERTopic and Top2Vec had significantly negative slopes (coefficients and p-values are reported in Appendix D).

Topeax also exhibited the lowest absolute percentage error in recovering the number of topics (see Figure 2) with $\text{MAPE} = 60.52$ ($\text{SD} = 26.19$), while Top2Vec ($M = 1797.29\%, \text{SD} = 2622.52$) and BERTopic ($M = 2438.91\%, \text{SD} = 3011.63$) drastically deviated from the number of gold labels in the datasets. It is also important to note the opposite directionality of these errors. While Topeax almost universally underestimated the number of topics, especially in `StackExchangeClusteringP2P` and `MedrxivClusteringP2P`, where the number of unique labels was very large, Top2Vec and BERTopic almost always grossly overestimated the number of clusters in the data. This is undesirable behaviour for a topic model, as topic interpretation requires manual effort, and vast numbers of topics (>500) become difficult and labour-intensive to label for any individual.

| Model | $C_{\text{in}}$ | $C_{\text{ex}}$ | $d$ | $I$ |
|---|---|---|---|---|
| Topeax | **0.35±0.15** | 0.32±0.09 | **0.96±0.05** | **0.55±0.10** |
| Top2Vec | 0.21±0.11 | **0.39±0.09** | 0.57±0.29 | 0.38±0.15 |
| BERTopic | 0.24±0.12 | 0.17±0.04 | 0.64±0.17 | 0.35±0.10 |

Table 1: Metrics of topic quality compared between different models. Best bold, second best underlined. Uncertainty is standard deviation. Higher is better.

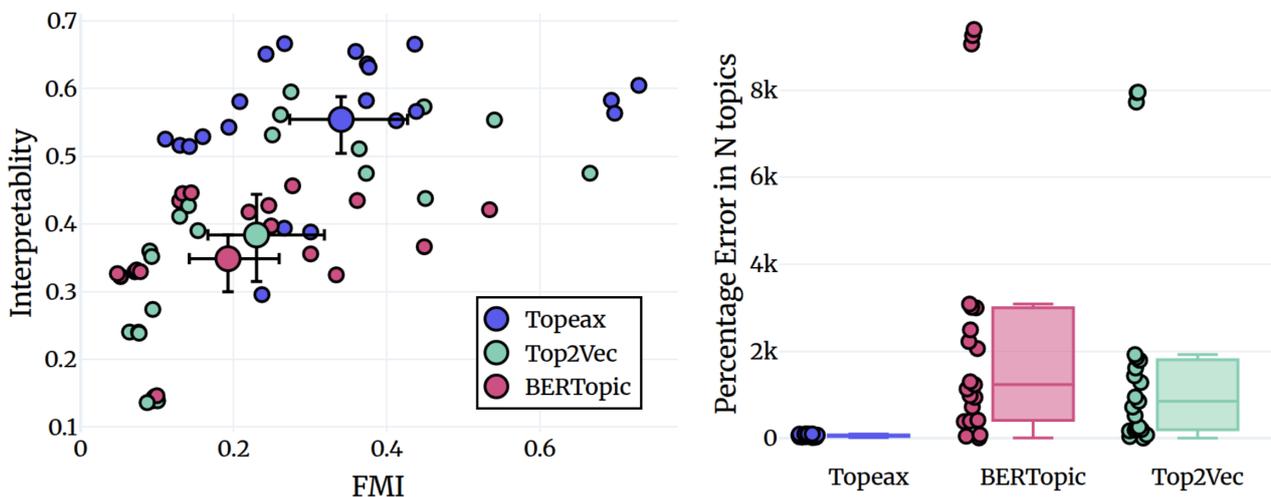

Figure 2: Performance comparison of clustering topic models.
*Left (Higher is better)*: Fowlkes-Mallows Index against topic interpretability. Large point with error bar represents mean with bootstrapped 95% confidence interval.
*Right (Lower is better)*: Distribution of absolute percentage error in finding the number of topics.

## 4.1. Perplexity

Metrics of quality and number of topics across perplexity values can are displayed on Figure 3. Topeax converges very early on the number of topics with perplexity, and remains stable from `perplexity=5`, while converges at around `perplexity=30` for quality metrics. It is reasonable to conclude that 50 is a good default value. Meanwhile, BERTopic converges at around `perplexity=50`, and has the lowest performance on all metrics. Top2Vec does not seem to converge at all in this range, and is most unstable. It does seem to improve with larger values of the hyperparameter. Keep in mind, that while BERTopic and Top2Vec improve with higher values, their default is set at `perplexity=15`, which, in light of these evaluations, is suboptimal.

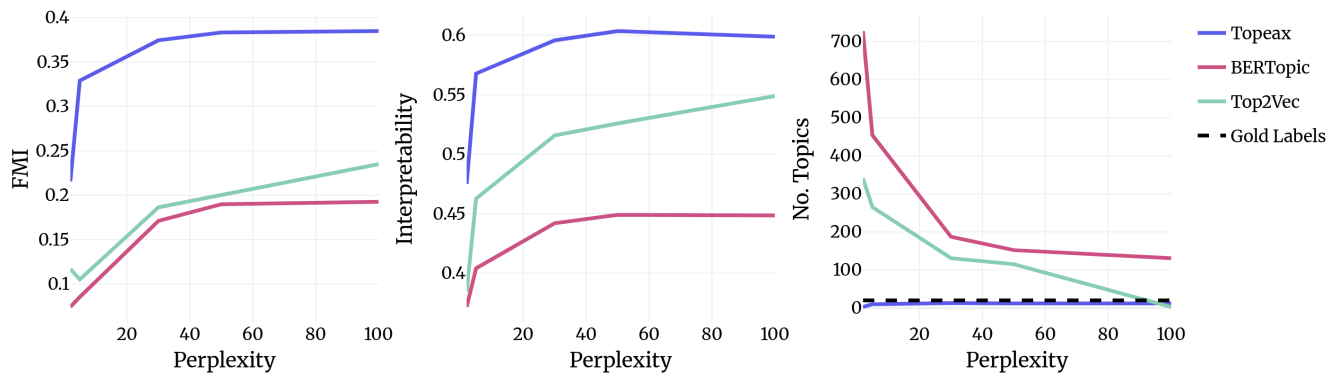

Figure 3: Clustering model's performance at different perplexity values.
*Left*: Fowlkes-Mallows Index at different perplexity values, *Middle*: Topic Interpretability Score at different values of Perplexity, *Right*: Number of Topics at each value of perplexity against Gold label.

### 4.2. Subsampling

Number of topics, topic quality and cluster quality are displayed on Figure 4. Topeax is relatively well-behaved, and converges to the highest performance when it has access to the full corpus. The number of topics is also relatively stable across from a sample size of 5000 (hovers around 10-12). In contrast, BERTopic and Top2Vec do not converge to a single value of N topics and keep growing with the size of the subsample. This also has an impact on cluster and topic quality. BERTopic has highest performance on the smallest subsamples (250-1000), while Top2Vec has best performance on a subsample of 5000, both methods decrease in performance as the number of topics grows with sample size. This behaviour is far from ideal, and it is apparent that Topeax is much more reliable at determining the number and structure of clusters.

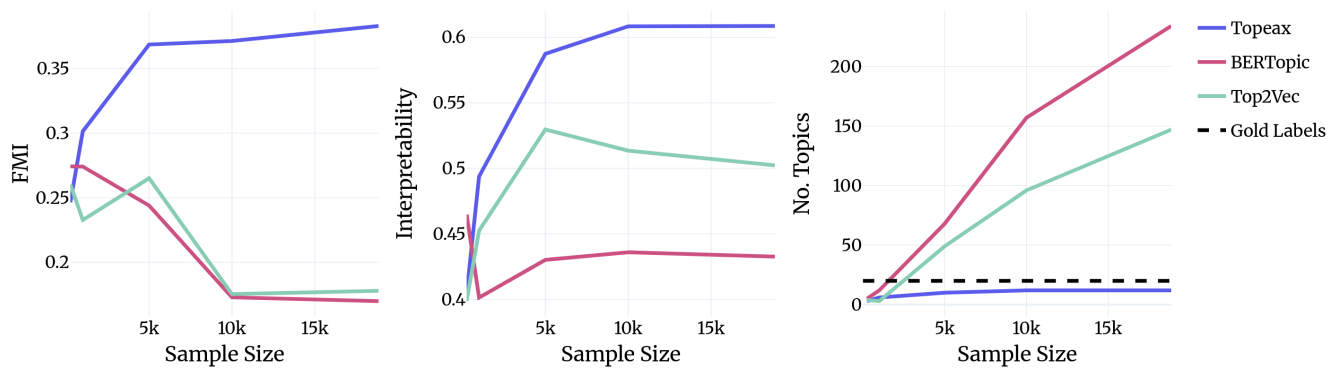

Figure 4: Topic models' performance at different subsample sizes.
*Left*: Fowlkes-Mallows Index as a function of sample size, *Middle*: Topic Interpretability Score at different subsamples, *Right*: Number of Topics discovered in each subsample per sample size.

### 4.3. Qualitative Considerations

As per the experimental evaluations presented above, Topeax systematically underestimates the number of clusters in a given dataset, despite matching the gold labels better. This warrants further investigation. A Topeax model was run on 20 Newsgroups with `all-MiniLM-L6-v2` embeddings, where the estimated number of clusters was 11, while the original dataset contains data from 20 categories. Adjusted mutual information was calculated between each topic discovered by the model and each newsgroup (see Figure 5).

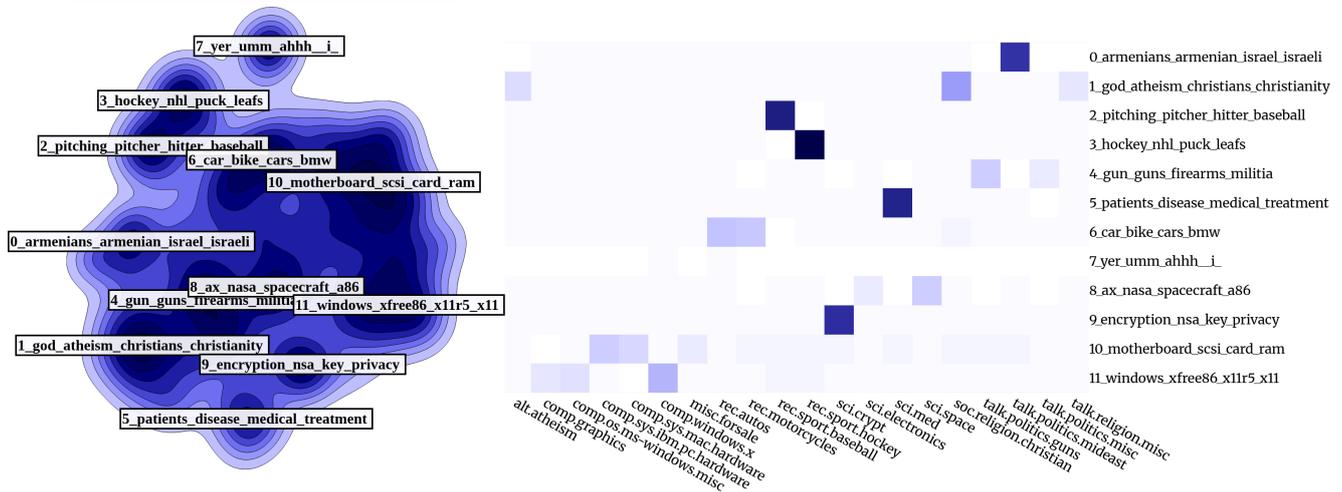

Figure 5: Topeax model fit on the 20 Newsgroups Corpus in relation to the gold labels provided in the corpus.
*Left*: Density estimate and density peaks annotated with top 4 keywords from each topic.
*Right*: Adjusted Mutual Information between cluster labels in the model, and gold labels in the corpus.

While, indeed the number of clusters is less than the categories in the original dataset, the clustering provided by Topeax is arguably just as natural. Most clusters ended up compressing information from one or two newsgroups, that were in some way related. For instance the `1_god_atheism_christians_christianity` topic contained documents from `alt.atheism`, `talk.religion.misc` and `soc.religion.christian`, thereby combining discourse on religion into a single topic. Likewise `6_car_bikes_bmw` compresses the `rec.autos` and `rec.motorcycles` newsgroups. In addition, the model uncovered a topic of outlier documents (`7_yer_umm_ahh__i_`), which were either empty, or contained no coherent sentences.

BERTopic discovered 232, and Top2Vec 145 topics in the same corpus using the same embeddings, while labelling 34.15% and 35.07% of documents as outliers respectively. While users might have different tolerance levels for time spent on analyzing topics, and the number of outliers, this behaviour seems far from ideal under most circumstances. Interpreting, and labelling the topics would take a considerable amount of time in both cases. In addition, regarding more than a third of documents as outliers means that a substantial amount of information is not covered by these models. This will inevitably prompt users of these topic models to a) hierarchically reduce topics, where they are required to specify the number of topics or b) change hyperparameters until they arrive at a result they deem sensible. It is thus questionable, whether these models are at all able to identify the number of natural clusters in a corpus, and until better and more rigorous heuristics are established for hyperparameter selection, their use remains highly subjective.

## 5. Conclusion

I propose a novel method, Topeax for finding natural clusters in text data, and assigning keywords to these clusters based on peak finding in kernel-density estimates. The model is compared to popular clustering topic models, Top2Vec and BERTopic on clustering datasets from the Massive Text Embedding Benchmark. In addition, models' robustness and stability to sample size and hyperparameter choices is evaluated. Topeax approximates human clusterings significantly better than previous approaches and describes topics with more diverse and coherent keywords. Furthermore, the model exhibits much more stable behaviour under changing sample size and hyperparameters. It is important to note, however, that Topeax underestimates the number of clusters systematically. Qualitative investigation suggests that this is due to the model grouping together related clusters in the case of 20 Newsgroups. In light of these findings, Topeax seems a better choice for text clustering,

## 6. Limitations

While the model has been shown to perform better than the baselines discussed, there are a number of issues it still exhibits:

1. Topeax underestimates the number of clusters, compared to humans.
2. The model, as of now, cannot be used in an online setting, when new topics have are as new information comes in.

Some of these issues might be addressed by using emerging dimensionality reduction techniques that allow for aligning between multiple datasets, and projection of out-of-distribution points. These issues should be subject to further investigation.

In addition the evaluation methodology also has a number of limitations of its own:

1. Quantitative metrics of topic quality, while roughly correlate with human preference, do not perfectly capture interpretability. Preferably, future research should evaluate topic quality with human subjects.
2. Subsampling and perplexity were only tested on the 20NG corpus in the interest of time and compute. This is of course a limitation, and evaluation on multiple corpora would be preferable.

# Appendix

# A Example code

Due to the model being implemented in Turftopic, you can easily run it on a corpus and print and plot the fitted model's results:

```
# pip install turftopic, datasets, plotly
from datasets import load_dataset
from turftopic import Topeax

ds = load_dataset("gopalkalpande/bbc-news-summary", split="train")
topeax = Topeax(random_state=42)
doc_topic = topeax.fit_transform(list(ds["Summaries"]))

topeax.plot_steps()
```

(see Figure 6)

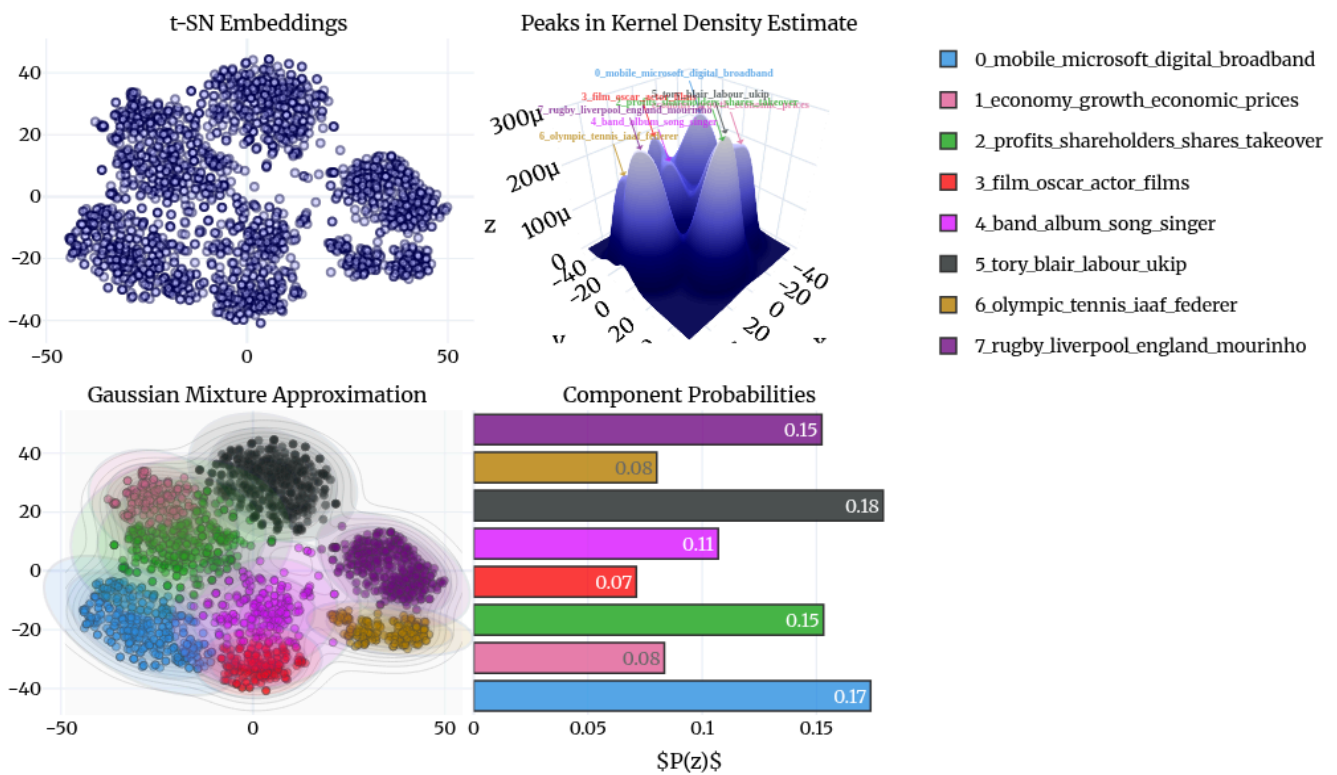

Figure 6: Interactive plot of steps in the Topeax algorithm on the BBC News dataset

```
topeax.print_topics()
```

(see Table 2)

| ID | Highest Ranking |
|---|---|
| 0 | mobile, microsoft, digital, technology, broadband, phones, devices, internet, mobiles, computer |
| 1 | economy, growth, economic, deficit, prices, gdp, inflation, currency, rates, exports |
| 2 | profits, shareholders, shares, takeover, shareholder, company, profit, merger, investors, financial |
| 3 | film, actor, oscar, films, actress, oscars, bafta, movie, awards, actors |
| 4 | band, album, song, singer, concert, rock, songs, rapper, rap, grammy |
| 5 | tory, blair, labour, ukip, mps, minister, election, tories, mr, ministers |
| 6 | olympic, tennis, iaaf, federer, wimbledon, doping, roddick, champion, athletics, olympics |
| 7 | rugby, liverpool, england, mourinho, chelsea, premiership, arsenal, gerrard, hodgson, gareth |

Table 2: Top 10 Keywords for the topics found in the BBC News corpus

# B C-TF-IDF

This section contains the formula for computing C-TF-IDF term importance.

- Let $C_{ij}$ be the number of times word j occurs in document i.
- $\text{tf}_{kj} = \frac{c_{kj}}{w_k}$, where $c_{kj} = \sum_{i \in k} C_{ij}$ is the number of occurrences of a word in a topic and $w_k = \sum_j c_{kj}$ is all words in the topic
- Estimate inverse document/topic frequency for term $j$: $\text{idf}_j = \log\left(1 + \frac{A}{\sum_k |c_{kj}|}\right)$, where $A = \frac{\sum_k \sum_j c_{kj}}{M}$ is the average number of words per topic, and $M$ is the number of topics.
- Calculate importance of term $j$ for topic $k$: $\beta_{kj} = \text{tf}_{kj} \cdot \text{idf}_j$

# C Descriptive Statistics for Datasets

Testing dataset statistics are reported in Table 3.

| Dataset | Document Length | Corpus Size | Clusters |
|---|---|---|---|
| | N characters | N documents | N unique gold labels |
| ArXivHierarchicalClusteringP2P | 1008.44±438.01 | 2048 | 23 |
| BiorxivClusteringP2P.v2 | 1663.97±541.93 | 53787 | 26 |
| MedrxivClusteringP2P.v2 | 1981.20±922.01 | 37500 | 51 |
| StackExchangeClusteringP2P.v2 | 1091.06±808.88 | 74914 | 524 |
| TwentyNewsgroupsClustering.v2 | 32.04±14.60 | 59545 | 20 |
| TweetTopicClustering | 165.66±68.19 | 4374 | 6 |
| BBCNewsClustering | 1000.46±638.41 | 2224 | 5 |

Table 3: Descriptive statistics of the datasets used for evaluation
*Document length is reported as mean±standard deviation*

# D Regression modelling

Coefficients for the model prediction FMI from model type are reported in Table 4.

| Coefficients | Estimate | p-value | 95% CI |
| --- | --- | --- | --- |
| Intercept (*Topeax*) | 0.3405 | 0.000 | [0.267, 0.414] |
| Topeax | −0.1106 | 0.038 | [−0.215, −0.006] |
| BERTopic | −0.1479 | 0.006 | [−0.252, −0.044] |

Table 4: Regression coefficients for predicting Fowlkes-Mallows Index from choice of topic model